\pdfoutput=1

\documentclass[11pt]{article}

\usepackage[]{naacl2021}

\usepackage{times}
\usepackage{latexsym}

\usepackage[T1]{fontenc}

\usepackage[utf8]{inputenc}

\usepackage{microtype}

\usepackage{graphicx}
\usepackage{subfigure}
\usepackage{hyperref}
\usepackage{booktabs} 
\graphicspath{ {images/} }
\usepackage{amsmath}

\title{MixUp Training Leads to Reduced Overfitting and Improved Calibration for the Transformer Architecture}

\author{Wancong Zhang \\
  Courant Institute of \\
  Mathematical Sciences \\
  New York University \\
  \texttt{wz1232@nyu.edu} \\\And
  Ieshan Vaidya \\
  Center for Data Science \\
  New York University \\
  \texttt{iav225@nyu.edu} \\}

\date{}

\begin{document}
\maketitle
\begin{abstract}
MixUp is a computer vision data augmentation technique that uses convex interpolations of input data and their labels to enhance model generalization during training. However, the application of MixUp to the natural language understanding (NLU) domain has been limited, due to the difficulty of interpolating text directly in the input space. In this study, we propose MixUp methods at the Input, Manifold, and sentence embedding levels for the transformer architecture, and apply them to finetune the BERT model for a diverse set of NLU tasks. We find that MixUp can improve model performance, as well as reduce test loss and model calibration error by up to $50\%$.

\end{abstract}

\section{Introduction}
MixUp is a data augmentation technique originated from computer vision, where convex interpolations of input images and their corresponding labels are used as additional training source. It has empirically been shown to improve the performance of image classifiers, and leads to decision boundaries that transition linearly between classes, providing a smoother estimate of uncertainty ~\citep{Zhang:18}.

Adapting MixUp to language is non-trivial and not well explored. Unlike with images, one can not interpolate texts in the input space directly. Several studies get around this issue by interpolating sentence or word embeddings, and show that MixUp can improve model performance for sentence classification \citep {Guo:19, Sun:20}.

We investigate whether MixUp can be extended to a wider range of NLU tasks including sentiment classification, multiclass classification, sentence acceptability, natural language inference (NLI), and question answering (QA). In addition, we propose a more diverse set of MixUp methods at the 1) Input, 2) Manifold, and 3) sentence-embedding levels for BERT \citep{Devlin:19}, and compare their performance for different task types at different resource settings. At last, we study the effect of MixUp on BERT's calibration, which tells us about the quality of a model's predictive uncertainty \citep{Guo:17}.

We find that 1) MixUp improves model performance over baselines for IMDb and AGNews, particularly at lower resource settings. 2) MixUp can significantly reduce the test loss and calibration error of BERT by up to $50\%$ without sacrificing performance. 3) While MixUp of sentence embeddings can be universally applied to a diverse set of NLU tasks, MixUp methods that interpolate individual tokens, such as Input and Manifold MixUp, are more appropriate for simpler tasks that do not necessarily require syntactic knowledge (eg. sentiment analysis).

\section{Background}
\subsection{MixUp}
In MixUp, convex interpolations of training input and their labels are used instead of the original data for training, with the effect of improved generalization and robustness against adversarial attacks. The following describes the MixUp process:

\begin{equation}\label{eq:1}
    \begin{gathered}
        \widetilde{x} = \lambda x_i + (1-\lambda)x_j \\
        \widetilde{y} = \lambda y_i + (1-\lambda)y_j
    \end{gathered}
\end{equation}

where $x_i, x_j$ are the input vectors, $y_i, y_j$ their one-hot coded labels, and $\lambda \in [0,1]$ is sampled from the Beta distribution: $\lambda \sim \beta(\alpha, \alpha)$ during every mini-batch, where $\alpha$ is a hyperparameter. Lower $\alpha$ leads to more even proportions between $\lambda$ and $1-\lambda$.

The application of MixUp can also be extended to the hidden representations. This is referred to as Manifold MixUp \citep{Verma:19}.

\subsection{MixUp for Language}
\citet{Guo:19} investigate two strategies to apply MixUp to language using an RNN architecture: 1) sentence-level MixUp interpolates the sentence embeddings of two sentences, while 2) word-level MixUp interpolate individual word embeddings at each position between two sentences. Both strategies have been shown to improve accuracy and regularize the training for simple sentence classification tasks such as sentiment analysis.

In a similar work done concurrently at the time of our experiments, \citet{Sun:20} applied sentence embedding MixUp to finetune transformers on GLUE tasks \citep{Glue}, and reported improved predictive performance for most tasks.

While \citep{Sun:20} focused on sentence embedding MixUp as the augmentation strategy and predictive performance as the evaluation metric, we evaluated a more diverse set of MixUp methods, including those operating on the input and manifold levels, on additional metrics including negative log-likelihood and model calibration. Unlike the previous work, we did not find significant predictive improvement for the shared GLUE tasks. On the other hand, we showed that MixUp can significantly improve model calibration and reduce overfitting for transformers, and that Input and Manifold MixUp methods work better than sentence embedding MixUp for content-based tasks.

\subsection{Model Calibration}
As neural networks become more widely used in high risk fields such as medical diagnosis and autonomous vehicles, the quality of their predictive uncertainty becomes an important feature. Models should not only be accurate, but also know when they are likely to be wrong.

To quantitatively evaluate a model's predictive uncertainty, \citet{Guo:17} defines calibration as the degree to which a model's predictive scores are indicative of the actual likelihood of correctness. \citet{Thulasidasan:19} show that MixUp can improve calibration for CNN based models. \citet{Desai:20} evaluate the calibration of pretrained transformers and finds it to substantially deteriorate under data shift.

\section{Methods}

\subsection{Architecture}
We used the \texttt{bert-base-uncased} pretrained model based on Hugging Face's implementation \citep{Wolf2019HuggingFacesTS}.

\subsection{Training}

Baseline models are trained using Empirical Risk Minimization (no MixUp) with cross entropy loss. Hyperparameters are given in Appendix \ref{sec:hyper}. 

MixUp models are trained using the MixUp methods discussed below, with otherwise identical architecture, hyperparameters, and training dataset (prior to augmentation).

Results are reported for the epoch with the best performance metric (accuracy or MCC).

\subsection{Datasets}
We choose to test on a diverse set of tasks in terms of both difficulties and task types. In particular:
\begin{itemize}
    \item \textbf{IMDb}: Sentiment classification for positive and negative movie reviews ~\citep{IMDb}
    \item \textbf{AGNews}: Multi-class classification for 4 topics of News articles. ~\citep{AGNews}
    \item \textbf{CoLA}: Classification for grammatically correct/incorrect sentences ~\citep{Glue}
    \item \textbf{RTE}: Natural language inference task, with entailment and no-entailment classes. ~\citep{Glue}
    \item \textbf{BoolQ}: Question-answering task that matches a short passage with a yes / no question about the passage. \citep{SuperGLUE}
\end{itemize}
\begin{table}[h!]
\small
\centering
\begin{tabular}{llrrr}
\toprule \textbf{Task} & \textbf{Metric} & \textbf{Train} & \textbf{Dev} & \textbf{Test} \\
\midrule
IMDb & Accuracy & 25000 & 3200 & 21800 \\
AGNews & Accuracy & 120000  & 2400 & 5200 \\
CoLA & Matt Corr & 8551 & 1043 & 1064 \\
RTE & Accuracy & 2491 & 278 & 3000 \\
BoolQ & Accuracy & 9427 & 3270 & 3245 \\
\bottomrule
\end{tabular}
\caption{\label{font-table} Datasets sizes (unit: sentence)}
\label{table:datasets}
\end{table}

We simulated low resource settings for IMDb and AGNews by training only using a random sample of 32 training examples.

\subsection{Metrics}
We evaluate our models based on 1) Accuracy, 2) Cross Entropy Loss, and 3) Expected Calibration Error (ECE).

Classifiers capable of reliably forecasting their accuracy are considered to be well-calibrated. For instance, a calibrated classifier should be correct $80\%$ of the time on examples to which it assigns $80\%$ confidence (the winning probability). Let the classifier's confidence for its prediction $\hat{Y}$ be $C$. Then, the Expected Calibration Error is the expectation of the absolute difference between the accuracy at a given confidence level and the actual confidence level:
\begin{equation}\label{eq:2}
    \begin{gathered}
        \mathrm{E_C}[|P(Y=\hat{Y}|C=c) - c|]
    \end{gathered}
\end{equation}
Details on how to empirically estimate this quantity can be found in Appendix \ref{sec:calib}.

\subsection{MixUp for Transformers}
We investigate three variants of MixUp that can be applied to the transformer architecture, in particular BERT: 1) CLS MixUp, 2) input-token MixUp, and 3) Manifold MixUp. As we shall see, the last is a generalization of the former two.

\subsubsection{CLS MixUp}
Unlike in computer vision, where inputs are preprocessed to the same dimensions, it is not straightforward to interpolate texts in the input space directly. Instead, we can perform MixUp on the sentence embeddings, which in the specific case of BERT are represented by the pooled CLS tokens from the final layer. Formally, given a pair of training input sentences $x_1, x_2$, their one hot label vectors $y_1, y_2$, and the sentence embeddings $f(x_1), f(x_2)$ generated using the BERT encoder, we interpolate their embeddings and the labels:
\begin{equation}\label{eq:3}
    \begin{gathered}
        \widetilde{x} = \lambda\: f(x_1) + (1-\lambda)\:f(x_2) \\
        \widetilde{y} = \lambda y_1 + (1-\lambda)y_2 \\
        \widetilde{y_p} = \mathrm{Classifier}(\widetilde{x})
    \end{gathered}
\end{equation}
MixUp samples, in this case embeddings and labels, are generated by interpolating each sample in the batch with another randomly selected sample. We feed the interpolated embeddings into the classifier (while discarding the unmixed embeddings, thus maintaining the same batch size) and minimize the cross entropy loss between the predictions $\tilde{y_p}$ and the interpolated labels $\tilde{y}$ in a given batch by summing their losses and backpropagating it through the entire computational graph.

We repeat this process stochastically for every batch, where a new $\lambda$ is sampled every time (Eq \ref{eq:1}).

\subsubsection{Input MixUp}
Though less intuitive, we can also interpolate individual input token embeddings at each index between two sentences.

Given sentences $x_1=\left[x_1^1, x_1^2, \cdots, x_1^m \right]$, $x_2 = \left[x_2^1, x_2^2, \cdots, x_2^n\right]$ and their one hot label vectors $y_1$ and $y_2$, where $m \geq n$ and $x^i$ represents the $i^{\mathrm{th}}$ input token embedding, we pad $x_1$ and $x_2$ to the same length using the \texttt{SEP} token from BERT vocabulary (see Appendix \ref{sec:padding} for an evaluation of different padding options). We can then generate a new sentence $\widetilde{x}$ by taking a MixUp of individual input token embeddings from each position:
\begin{equation}\label{eq:4}
    \begin{gathered}
        \widetilde{x}^i = \lambda x_1^i + (1 - \lambda) x_2^i \ ; \ i \in (1,...,m) \\
        \widetilde{x} = [\widetilde{x}^1, \widetilde{x}^2, ..., \widetilde{x}^m] \\
        \widetilde{y} = \lambda y_1 + (1-\lambda)y_2 \\
        \widetilde{y_p} = \mathrm{Classifier}(f(\widetilde{x}))
    \end{gathered}
\end{equation}
Where $f(x)$ is the sentence embedding returned by the BERT encoder given input tokens $x$.

\subsubsection{Manifold MixUp}
\citet{Verma:19} demonstrated that applying MixUp to the hidden represenations, in addition to the inputs, yields additional benefit in reducing test errors in computer vision. Given that BERT consists of multiple layers, we also investigate the effect of this technique on its training.

\begin{figure}[t]
   \centering
   \includegraphics[width=\columnwidth]{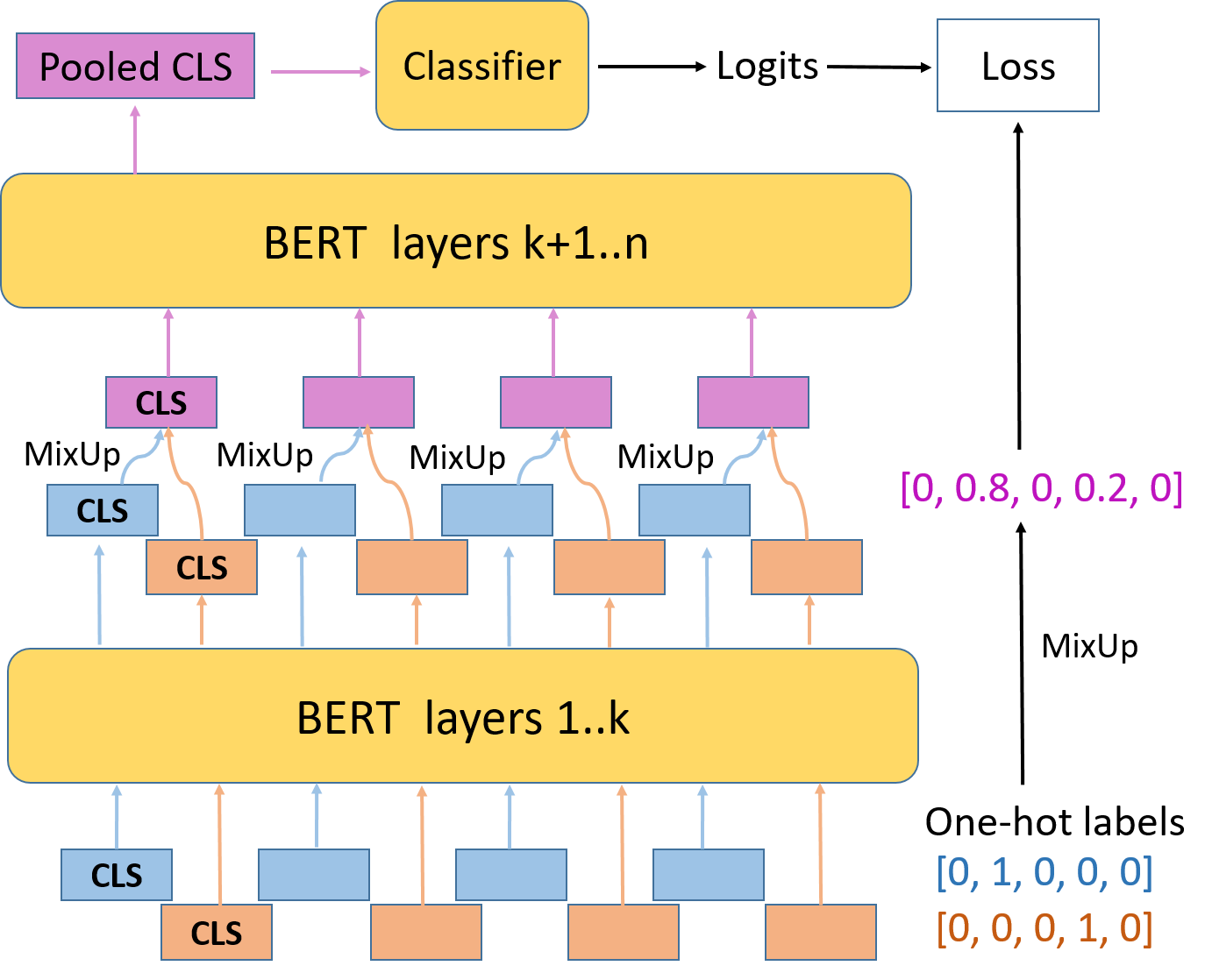}
   
   \caption{Schematics of Manifold MixUp. We mix the layer $k$ embeddings and the one-hot labels of two training samples using the same proportions $\lambda$ and $(1-\lambda)$ (See eq 1). }
   \label{fig:manifold}
\end{figure}

We proceed as follows: for each batch, select a random layer $k$ from a set of eligible layers $S$ in the BERT with $n$ encoder layers. This set may include the input layer $0$ as well as the pooled sentence embedding layer $n+1$. We apply MixUp on the hidden representation embeddings, before feeding the interpolated embeddings to the next layer to continue with training. The loss is computed using the logits and the interpolated labels (Figure \ref{fig:manifold}). 

When $S$ only contains $0$, Manifold MixUp reduces to Input MixUp. When $S$ only contains $n+1$, it reduces to CLS MixUp.

\section{Results}
Our IMDB (Table \ref{table:1}) and AGNews (Table \ref{table:2}) results indicate that Manifold, Input, and CLS MixUp methods improve model robustness and calibration across a range of data regimes; the improvements are more significant when the amount of training data is scarce, which matches our expectation since MixUp can be seen as a form of data augmentation. 
Manifold MixUp offers the best improvement for model calibration; in fact, the Manifold MixUp model trained on merely 32 samples is equally or more calibrated in comparison to the baseline model trained using the full training set. 

Both sentiment analysis on IMDB and news category classification on AGNews are relatively simple tasks that do not necessarily require the model to have syntactic knowledge; instead, they are likely to be solved by only inspecting the sentiment or classification of the individual words. For example, a movie review with more negative words than positive ones is likely to be negative overall. Therefore, we expect Input and Manifold MixUp, which interpolate embeddings at the token level, to work well here; indeed, they outperform baseline and CLS MixUp in these simpler scenarios.

\begin{table}[t]
\small
\centering
\begin{tabular}{llrcr}
\toprule \textbf{Size} & \textbf{Model} & \textbf{Acc} & \textbf{Loss} & \textbf{ECE} \\
\midrule
Full & Base & $91.8 \pm 0.1$ & $61 \pm 8$ & $7.7 \pm 0.1$ \\
 & CLS & $92.0 \pm 0.1$  & $29 \pm 1$ & $5.8 \pm 0.2$ \\
 & Input & $92.1 \pm 0.1$ & $37 \pm 2$ & $6.8 \pm 0.2$ \\
 & Mani & $91.8 \pm 0.1$ & $27 \pm 1$ & $4.4 \pm 0.6$ \\
\midrule
$32$ & Base & $68.2 \pm 1.6$ & $180 \pm 19$ & $29.0 \pm 1.5$ \\
 & CLS & $69.6 \pm 1.7$  & $86 \pm 4$ & $21.0 \pm 2.8$ \\
 & Input & $70.8 \pm 1.2$ & $61 \pm 7$ & $8.9 \pm 2.3$ \\
 & Mani & $72.3 \pm 1.0$ & $55 \pm 2$ & $3.4 \pm 0.9$ \\
\bottomrule
\end{tabular}
\caption{\label{font-table} MixUp experiments for IMDB using various models (baseline, cls, input, manifold), train sizes. Standard errors over five runs. }
\label{table:1}
\end{table}

\begin{table}[t]
\small
\centering
\begin{tabular}{llrcr}
\toprule \textbf{Size} & \textbf{Model} & \textbf{Acc} & \textbf{Loss} & \textbf{ECE} \\
\midrule
Full & Base & $93.7 \pm 0.1$ & $36 \pm 7$ & $5.0 \pm 0.6$ \\
 & CLS & $93.7 \pm 0.1$  & $28 \pm 1$ & $4.3 \pm 0.3$ \\
 & Input & $93.8 \pm 0.1$ & $25 \pm 1$ & $4.1 \pm 0.3$ \\
 & Man & $93.7 \pm 0.1$ & $23 \pm 5$ & $3.0 \pm 0.2$ \\
\midrule
$32$ & Base & $81.9 \pm 0.6$ & $72 \pm 9$ & $9.9 \pm 1.9$ \\
 & CLS & $83.4 \pm 0.6$  & $55 \pm 2$ & $7.2 \pm 0.7$ \\
 & Input & $82.9 \pm 0.3$ & $53 \pm 2$ & $6.2 \pm 2.4$ \\
 & Mani & $82.3 \pm 0.4$ & $52 \pm 1$ & $4.8 \pm 1.8$ \\
\bottomrule
\end{tabular}
\caption{\label{font-table} MixUp experiments for AGNews using various train sizes. Standard errors over five runs. }
\label{table:2}
\end{table}

Unlike IMDB and AGNews, the remaining tasks CoLA (acceptability), RTE (NLI), and BoolQ (QA) rely on syntactic knowledge. Therefore, we would expect Input and Manifold MixUp to work relatively poorly for these tasks, since they can disrupt sentence syntax (eg. by interpolating nouns with verbs). This is empirically confirmed when we applied the three MixUp methods to training on the CoLA dataset (Table \ref{table:3}), where the Input and Manifold MixUp models under-perform baseline.

\begin{table}[t]
\small
\centering
\begin{tabular}{llrcr}
\toprule \textbf{Task} & \textbf{Model} & \textbf{MCC/Acc} & \textbf{Loss} & \textbf{ECE} \\
\midrule
CoLA & Base & $52.5 \pm 0.6$ & $56 \pm 4$ & $11.5 \pm 0.3$ \\
& CLS & $52.8 \pm 0.6$  & $48 \pm 1$ & $9.7 \pm 0.1$ \\
& Input & $50.8 \pm 0.8$  & $48 \pm 8$ & $10.1 \pm 1.4$ \\
& Mani & $51.0 \pm 1.0$  & $45 \pm 2$ & $9.0 \pm 0.9$ \\
\midrule
RTE & Base & $63.8 \pm 1.4$ & $125 \pm 20$ & $27.6 \pm 1.9$ \\
 & CLS & $63.8 \pm 1.0$  & $80 \pm 4$ & $20.8 \pm 1.3$ \\
\midrule
BoolQ & Base & $73.4 \pm 0.3$ & $170 \pm 7$ & $24.8 \pm 0.3$ \\
 & CLS & $73.1 \pm 0.3$  & $89 \pm 1$ & $21.8 \pm 0.3$ \\

\hline
\end{tabular}
\caption{\label{font-table} MixUp experiments for CoLA, RTE, and BoolQ. CoLA is evaluated using Test Matthew Correlation Coefficient (MCC), others using Test Accuracy. Standard errors over five runs.}
\label{table:3}
\end{table}

For the three more advanced tasks, while CLS MixUp exerts negligible effect on the model's predictive performance, it still significantly reduces test loss and improves calibration without sacrificing performance. The evolution of test losses across training steps for MixUp and baseline models are plotted in Appendix \ref{sec:loss}, demonstrating the ability of MixUp to reduce model overfitting. We note that in contrast to our experiments, \citet{Sun:20} were able to obtain improvements using CLS MixUp for CoLA and RTE using a different training regimen, where MixUp is only applied during the last half of training epochs.

\section{Conclusion}
In this work, we propose CLS, Input, and Manifold MixUp methods for the transformer architecture and apply them to a range of NLU tasks including sentiment analysis, multi-class classification, acceptability, NLI, and QA. We find that MixUp can improve the predictive performance of simpler tasks. More generally, MixUp can significantly reduce model overfitting and improve model calibration.

\bibliography{naacl2021}

\begin{thebibliography}{13}
\expandafter\ifx\csname natexlab\endcsname\relax\def\natexlab#1{#1}\fi

\bibitem[{Desai and Durrett(2020)}]{Desai:20}
Shrey Desai and Greg Durrett. 2020.
\newblock \href {https://arxiv.org/pdf/2003.07892.pdf} {Calibration of
  pre-trained transformers}.
\newblock arXiv:3003.07892.

\bibitem[{Devlin et~al.(2019)Devlin, Chang, Lee, and Toutanova}]{Devlin:19}
Jacob Devlin, Ming-Wei Chang, Kenton Lee, and Kristina Toutanova. 2019.
\newblock \href {https://arxiv.org/pdf/1810.04805.pdf} {Bert: Pre-training of
  deep bidirectional transformers for language understanding}.
\newblock In \emph{Proceedings of NAACL-HLT}.

\bibitem[{Guo et~al.(2017)Guo, Pleiss, Sun, and Weinberger}]{Guo:17}
Chuan Guo, Geoff Pleiss, Yu~Sun, and Kilian~Q. Weinberger. 2017.
\newblock \href {https://arxiv.org/pdf/1706.04599.pdf} {On calibration of
  modern neural networks}.
\newblock In \emph{Proceedings of the 34th International Conference on Machine
  Learning}.

\bibitem[{Guo et~al.(2019)Guo, Mao, and Zhang}]{Guo:19}
Hongyu Guo, Yongyi Mao, and Richong Zhang. 2019.
\newblock \href {https://arxiv.org/pdf/1905.08941.pdf} {Augmenting data with
  mixup for sentence classification: An empirical study}.
\newblock arXiv:1905.08941.

\bibitem[{Maas et~al.(2011)Maas, Daly, Pham, Huang, Ng, and Potts}]{IMDb}
Andrew~L. Maas, Raymond~E. Daly, Peter~T. Pham, Dan Huang, Andrew~Y. Ng, and
  Christopher Potts. 2011.
\newblock \href {http://www.aclweb.org/anthology/P11-1015} {Learning word
  vectors for sentiment analysis}.
\newblock In \emph{Proceedings of the 49th Annual Meeting of the Association
  for Computational Linguistics: Human Language Technologies}, pages 142--150,
  Portland, Oregon, USA. Association for Computational Linguistics.

\bibitem[{Sun et~al.(2020)Sun, Xia, Yin, Liang, Yu, and He}]{Sun:20}
Lichao Sun, Congying Xia, Wenpeng Yin, Tingtin Liang, Philip~S. Yu, and Lifang
  He. 2020.
\newblock \href {https://arxiv.org/pdf/2010.02394.pdf} {Mixup-transformer:
  Dynamic data augmentation for nlp tasks}.
\newblock In \emph{The 28th International Conference on Computational
  Linguistics}.

\bibitem[{Thulasidasan et~al.(2019)Thulasidasan, Chennupati, Bilmes,
  Bhattacharya, and Michalak}]{Thulasidasan:19}
Sunil Thulasidasan, Gopinath Chennupati, Jeff Bilmes, Tanmoy Bhattacharya, and
  Sarah Michalak. 2019.
\newblock \href {https://arxiv.org/pdf/1905.11001.pdf} {On mixup training:
  Improved calibration and predictive uncertainty for deep neural networks}.
\newblock In \emph{33rd Conference on Neural Information Processing Systems}.

\bibitem[{Verma et~al.(2019)Verma, Lamb, Beckham, Najafi, Mitliagkas,
  Courville, Lopez-Paz, and Bengio}]{Verma:19}
Vikas Verma, Alex Lamb, Christopher Beckham, Amir Najafi, Ioannis Mitliagkas,
  Aaron Courville, David Lopez-Paz, and Yoshua Bengio. 2019.
\newblock \href {https://arxiv.org/pdf/1806.05236.pdf} {Manifold mixup: Better
  representations by interpolating hidden states}.
\newblock In \emph{Proceedings of the 36th International Conference on Machine
  Learning}.

\bibitem[{Wang et~al.(2019)Wang, Pruksachatkun, Nangia, Singh, Michael, Hill,
  Levy, and Bowman}]{SuperGLUE}
Alex Wang, Yada Pruksachatkun, Nikita Nangia, Amanpreet Singh, Julian Michael,
  Felix Hill, Omer Levy, and Samuel Bowman. 2019.
\newblock \href {https://arxiv.org/pdf/1905.00537.pdf} {Superglue: A stickier
  benchmark for general-purpose language understanding systems}.
\newblock \emph{33rd Conference on Neural Information Processing Systems}.

\bibitem[{Wang et~al.(2018)Wang, Singh, Michael, Hill, Levy, and Bowman}]{Glue}
Alex Wang, Amanpreet Singh, Julian Michael, Felix Hill, Omer Levy, and Samuel
  Bowman. 2018.
\newblock \href {https://doi.org/10.18653/v1/w18-5446} {Glue: A multi-task
  benchmark and analysis platform for natural language understanding}.
\newblock \emph{Proceedings of the 2018 EMNLP Workshop BlackboxNLP: Analyzing
  and Interpreting Neural Networks for NLP}.

\bibitem[{Wolf et~al.(2019)Wolf, Debut, Sanh, Chaumond, Delangue, Moi, Cistac,
  Rault, Louf, Funtowicz, and Brew}]{Wolf2019HuggingFacesTS}
Thomas Wolf, Lysandre Debut, Victor Sanh, Julien Chaumond, Clement Delangue,
  Anthony Moi, Pierric Cistac, Tim Rault, R'emi Louf, Morgan Funtowicz, and
  Jamie Brew. 2019.
\newblock \href {https://arxiv.org/pdf/1910.03771.pdf} {Huggingface's
  transformers: State-of-the-art natural language processing}.
\newblock \emph{ArXiv}, abs/1910.03771.

\bibitem[{Zhang et~al.(2018)Zhang, Cisse, Dauphin, and Lopez-Paz}]{Zhang:18}
Hongyi Zhang, Moustapha Cisse, Yann~N. Dauphin, and David Lopez-Paz. 2018.
\newblock \href {https://arxiv.org/pdf/1710.09412.pdf} {mixup: Beyond empirical
  risk minimization}.
\newblock In \emph{International Conference on Learning Representations}.

\bibitem[{Zhang et~al.(2015)Zhang, Zhao, and LeCun}]{AGNews}
Xiang Zhang, Junbo Zhao, and Yann LeCun. 2015.
\newblock \href {http://arxiv.org/abs/1509.01626} {Character-level
  convolutional networks for text classification}.

\end{thebibliography}
\bibliographystyle{naacl2021}

\appendix

\section{Calibration Metrics}
\label{sec:calib}
As in the work of \citet{Guo:17}, we calculate the expected calibration error (ECE) of our model as follows:
Predictions are grouped into $15$ interval bins of equal size ordered by prediction confidence. Let $B_m$ be the set of samples that fall into bin $m$. The accuracy and confidence of $B_m$ are defined as:

$$
\mathrm{acc}(B_m) = \frac{1}{|B_m|} \sum_{i \in B_m} 1(\hat{y_i} = y_i)
$$
$$
\mathrm{conf}(B_m) = \frac{1}{|B_m|} \sum_{i \in B_m} \hat{p_i}
$$
where $\hat{p_i}$ is the confidence (winning score) of sample $i$. The Expected Calibration Error (ECE) is then defined as:
$$
\mathrm{ECE} = \sum^M_{m=1} \frac{|B_m|}{n} \Big|\mathrm{acc}(B_m) - \mathrm{conf}(B_m)\Big|
$$

\section{Padding Options for Input and Manifold MixUp}
\label{sec:padding}
In the previous MixUp study in language \citep{Guo:19}, all sentences are padded to the same length prior to training using Input MixUp. We believe that while sentences should be padded to the same length, it ought to be done using the minimum amount of padding tokens possible, as the introduction of excessive padding tokens can disrupt performance. Instead of uniformly padding all sentences to length of the longest sentence, we propose to pad pairs of sentences about to be interpolated during training to the length of the longer sentence.

We consider \texttt{SEP}, \texttt{unused0}, \texttt{PAD} tokens from the BERT vocabulary to be potential padding candidates. We empirically evaluate the effect of using each of these paddings for Input MixUp, both when using our proposal (pair) and the one in which all sentences are padded to the length of the longest sequence (max).

We performed padding experiments on SST-2 dataset ~\citep{Glue} due to its short training sentence lengths, which leaves more room for padding. We simulate low resource setting by using a random sample of 64 training samples to increase task difficulty, thereby making it easier to detect performance differences.

Our results (Table \ref{table:paddings}) confirm that pair-wise padding works better than no padding at all as well as the uniform padding of sentences to the maximum length, while the choice of \texttt{SEP} results in the best performance.

\begin{table}[t]
\centering
\begin{tabular}{lr}
\toprule \textbf{Type} & \textbf{Accuracy} \\
\midrule
No Pad & $78.38 \pm 3.2$ \\
unused (pair) & $80.22 \pm 1.2$ \\
SEP (pair) & $\pmb{82.00 \pm 1.4}$ \\
PAD (pair) & $81.20 \pm 1.0$ \\
unused (max) & $68.14 \pm 5.3$ \\
SEP (max) & $69.21 \pm 5.6$ \\
PAD (max) & $73.61 \pm 3.3$ \\
\bottomrule
\end{tabular}
\caption{\label{font-table} The effect of different padding strategies on finetuning BERT on the SST-2 dataset using Input MixUp.}
\label{table:paddings}
\end{table}

\section{Hyperparameters}
\label{sec:hyper}

Hyperparameters for training downstream tasks are shown in Table \ref{table:hyperparameters}.

\begin{table}[t]
\small
\centering
\begin{tabular}{lcccccc}
\toprule \textbf{ } & \textbf{LR} & \textbf{BS} & \textbf{EP} & \textbf{DR} & \textbf{MSL} & \textbf{$\alpha$} \\
\midrule
IMDB & $1E-5$ & $16$ & $10$ & 0.1 & $128$ & $1.00$ \\
AGNews & $2E-5$ & $16$ & $10$ & 0.1 & $128$ & $1.00$ \\
CoLA & $2E-5$ & $16$ & $5$ & 0.1 & $128$ & $0.75$ \\
RTE & $3E-5$ & $16$ & $5$ & 0.1  & $128$ & $1.00$\\
BoolQ & $2E-5$ & $16$ & $10$ & 0.1 & $256$ & $1.00$\\
\bottomrule
\end{tabular}
\caption{\label{font-table} Hyperparameters for training downstream tasks. LR: Learning Rate. BS: Batch Size. EP: Training Epochs. DR: Dropout rate for BERT and classifier. MSL: Maximum Sequence Length. $\alpha$: hyperparameter for beta distribution used in  MixUp. }
\label{table:hyperparameters}
\end{table}

\section{Evolution of test loss}
\label{sec:loss}

\begin{figure}[t!]
   \centering
   \includegraphics[width=\columnwidth]{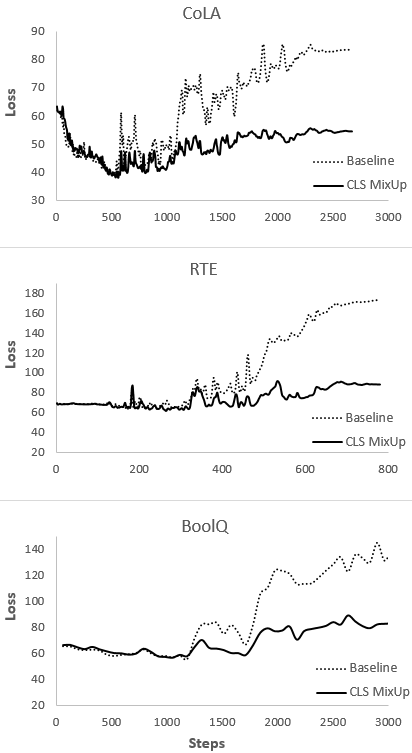}
   \caption{Test losses obtained by the baseline models versus CLS MixUp models across training steps}
   \label{fig:losses}
   \vspace{128in}
\end{figure}

Evolution of test losses across training steps for CoLA, RTE, and BoolQ (\ref{fig:losses}) are plotted. MixUp reduces overfitting by allowing the model to converge to a much lower test loss.

\end{document}